%% file: main.tex
\documentclass{article}

\usepackage[numbers, compress]{natbib}
\usepackage{caption}
\usepackage{subcaption}
\usepackage{graphicx}

\usepackage{adjustbox}
\usepackage{amsmath}
\usepackage[final, nonatbib]{tackling_climate_workshop_style}

\usepackage[utf8]{inputenc} % allow utf-8 input
\usepackage[T1]{fontenc}    % use 8-bit T1 fonts
\usepackage{hyperref}       % hyperlinks
\usepackage{url}            % simple URL typesetting
\usepackage{booktabs}       % professional-quality tables
\usepackage{amsfonts}       % blackboard math symbols
\usepackage{nicefrac}       % compact symbols for 1/2, etc.
\usepackage{microtype}      % microtypography

\title{CityTFT: Temporal Fusion Transformer for Urban Building Energy Modeling}

\author{%
  Ting-Yu Dai \\
  The University of Texas at Austin\\
  Austin, TX 78712 \\
  \texttt{funnyengineer@utexas.edu} \\
  \And
    Dev Niyogi \\
    The University of Texas at Austin \\
  Austin, TX 78712 \\
  \texttt{dev.niyogi@jsg.utexas.edu} \\
  \And
  Zoltan Nagy \\
  The University of Texas at Austin\\
  Austin, TX 78712 \\
  \texttt{nagy@utexas.edu} \\
}
\begin{document}

\maketitle

\begin{abstract}
    Urban Building Energy Modeling (UBEM) is an emerging method to investigate urban design and energy systems against the increasing energy demand at urban and neighborhood levels. However, current UBEM methods are mostly physic-based and time-consuming in multiple climate change scenarios. This work proposes CityTFT, a data-driven UBEM framework, to accurately model the energy demands in urban environments. With the empowerment of the underlying TFT framework and an augmented loss function, CityTFT could predict heating and cooling triggers in unseen climate dynamics with an F1 score of 99.98 \% while RMSE of loads of 13.57 kWh.
\end{abstract}
\input{contents/intro}
\input{contents/method}
\input{contents/result}

\input{contents/conclusion}

\medskip
\bibliography{refs}

\appendix
\input{contents/appendix}
\end{document}

%% file: contents/intro.tex
\section{Introduction}

% telling 1. the urban population is still growing 2. large proportion of emissions come from the urban area -> it is important to building urban level emission relationship 
Urbanization is one of the greatest challenges of modern society. Almost one-third of global greenhouse gas emissions come from buildings and 70\% of energy is consumed by urban. As of the latest available data in 2021, the global population stood at over 7.8 billion individuals \cite{unitednations2021}. Projections indicate that in 2050 \cite{population_2022}, the world's population may surpass 9.7 billion, reflecting a substantial demographic expansion over the intervening years. To address the long-term challenge posed by urbanization effectively, urban building energy modeling emerges as an imperative and requisite approach in academic research and urban planning endeavors.

% The purpose of UBEM / review UBEM works
The fundamental purpose of urban building energy modeling is to simulate and analyze the intricate dynamics of energy consumption within urban environments. Compared to building energy modeling, UBEM simulates while considering building height, surface coverage, and spatial arrangement, probing their interactions and discerning their collective influence on energy dynamics at the urban scale. \citeauthor{swan2009modeling} \cite{swan2009modeling} set a tone in reviewing urban modeling of the residential sector, subdividing the modeling methodologies into top-down and bottom-up approaches. Specifically, bottom-up physic-based UBEM methods have garnered attention in recent times \cite{nouvel2015simstadt}. \citeauthor{robinson2009citysim} developed CitySim to assist urban settlements with sustainable planning and also simulate the energy use of a few buildings ranging from tens to thousands. Those UBEM methods are reasonably accurate in simulating the performance of almost any building combination systems \cite{hensen2012building}. However, customized urban projects, and optimization problems involving many UBEM evaluations, are time-consuming and labor-intensive, and based on the scalability, the simulation runtime can be exponentially high if a broad set of design variations is analyzed. 

Here, we propose a surrogate data-driven approach to accelerate the simulation process in UBEM. Compared to similar previous works \cite{westermann2020using, westermann2021using, vazquez2019deep}, we based our work on the extensively used forecasting model, Temporal Fusion Transformer (TFT) \cite{lim2021temporal}. We extract the static covariate encoder and variable selection network from the TFT structure while adding a small neural network to model the probability of triggering heating and cooling needs. The major improvements are: 1.) Sequential input has been applied in a transformer-based model to improve the temporal accuracy. 2.) A training strategy that models from weather dynamics and urban interactions to energy demands simultaneously with a customized loss function. 3.) Improved generalizability for the proposed surrogate model. Urban planners or Energy Sectors could benefit from urban-level energy demand estimations by the proposed approach to enhance the decision-making process.

% useless sentences
% In 2013, \citeauthor{nouvel2015simstadt} \cite{nouvel2015simstadt} was developed as an urban energy simulation platform to support the planning of the energy transition at the urban scale. 
% As of September 2021, the proportion of the global population residing in urban areas amounted to approximately 55\%, while projections foresee a notable escalation to around two-thirds in 2050.
% While building energy models focus primarily on individual structures and their energy performance, UBEM extends its scope to encompass entire urban morphology including trees and pavements.
% The principal objective of UBEM is to advance the efficiency and sustainability of urban landscapes through the elucidation of intricate energy consumption dynamics, enabling informed decision-making processes, and fostering the formulation of strategies for ameliorating energy efficiency and mitigating environmental impacts.
% Nevertheless, the prevailing trend persists unabated. 
% Urban Building Energy Modeling (UBEM) is a well-discussed topic in the past decade due to the net zero policy and global warming. 

%% file: contents/method.tex
\section{Methods}
\newcommand{\probP}{\text{I\kern-0.26em P}}
\begin{figure}
    \centering
    \includegraphics[width=\textwidth]{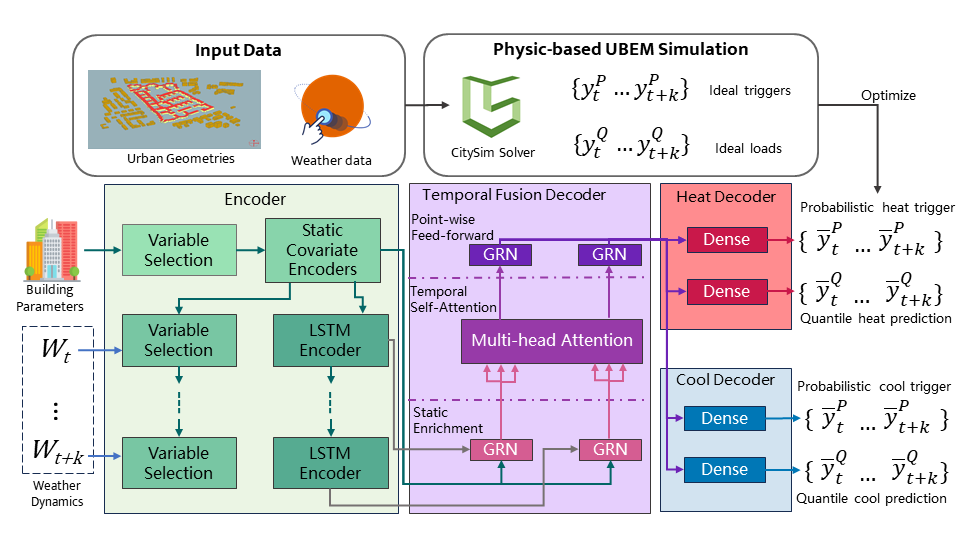}
    \caption{Overview of CityTFT. CitySim and CityTFT apply the same weather input while CityTFT uses simplified building parameters.}
    \label{fig:method}
\end{figure}
\textbf{Data Prepration} As a surrogate model, this work aims to duplicate the ability of physic-based UBEM models, CitySim \cite{robinson2009citysim}, to simulate urban building energy. Two major inputs of UBEM are urban geometries and weather data. This study utilized the university campus geometries and the calibrated building parameters from \cite{dilsiz2023spatio} which contains 114 buildings in a wide variety. 13 variables are extracted from the building geometries to represent the static building covariates. The detailed transformation method can be found in the appendix. The other requirement of CitySim is weather information. This study collects 21 Typical Meteorological Year (TMY) files in different climate zones globally by Meteonorm \cite{remund2020meteonorm} to obtain fruitful weather dynamics. 12 environmental variables are served in climate files as the temporal covariates while one additional variable, hour of year, is added to improve the temporal identification. Those variables are listed in Table \ref{tab:weather_vars}. The ideal hourly heating and cooling demands are simulated through CitySim solver for each building which concludes a dataset with roughly 17 million samples. The sequence length in our training process is 24. All the building variables, weather data, and electricity loads have been normalized for preprocessing.

\textbf{Temporal Fusion Transformer}
The Temporal Fusion Transformer (TFT) is composed of key elements: a variable selection network, static covariate encoder, and temporal processing, facilitating precise energy demand modeling. The variable section network is applied to both static building properties and temporal weather covariates to provide instance-wise variable selection. Linear transformations are applied to transform each variable into a $d_{model}$ dimension vector to match the subsequent paper for skip connections. Gated Residual Networks (GRN) are utilized for static enrichment and point-wise feed-forward processes while self-attention layers are utilized for temporal feature processing. To decode the transformed latent correctly, the decoder of TFT first applied GRN to enrich temporal signals with static building latents. Secondly, another self-attention network is applied to assemble static-enriched features into a single matrix. The attention mechanism here is to help that each temporal dimension can only attend to features preceding it. 

% improvement 
Different from the original TFT, this work aims to predict the energy demands in the same temporal period i.e. sequential modeling, unlike the purpose of the forecasting mission in the original paper. Therefore, this work aborts the future encoder part from the model and applies the original observed temporal features to proceed with the static enrichment and self-attention processing. The static enrichment is applied to embedded weather representations, and an interpretable attention mechanism builds the attention matrix in each time step of the enriched features. We also add one more linear layer with sigmoid activation for probabilistic projection. Heating and Cooling loads are divided into two variables while cooling loads are negative and heating loads are positive.

\textbf{Probabilistic Loss: Will it trigger and how much it will cost}
Since about 30 and 50 percent of the heating and cooling loads are zero, this work proposes a probabilistic-based loss to handle the imbalanced data. The loss equation is summarized in Eq. \ref{equ:overall} where $t$ is whether heating/cooling systems are triggered, $y^P$ is the probabilistic output produced by the TFT model, $a$ is the actual heating/cooling loads, and $y^Q$ is the quantile projections by the TFT model. Instead of computing deterministic energy consumption predictions, a probabilistic output of whether the heating/cooling system will be triggered is computed first. Eq. \ref{equ:prob} calculates the loss between the output and the triggering probability from target loads. On the other side, the quantile loss in Eq. \ref{equ:quant} only optimizes with the projections that are paired with the non-zero loads while $q$ represents the target quantile. The quantile loss aims to optimize the range of likely target values instead of a deterministic output, and the probability loss focuses on preventing our networks from predicting smaller values while optimizing with void loads. 
\begin{equation} \label{equ:overall}
    l = l_{prob}(t, y^P) + l_{quantile}(a, y^Q)
\end{equation}
\begin{equation} \label{equ:prob}
    l_{prob}(t, y^P) = \frac{1}{N} \sum_{n=1}^N \{ -w_n{(y_{n}^P \cdot \log(t_{n})}) + (1 - y_{n}^P) \cdot \log(1 - t_{n})) \}
\end{equation}

\begin{equation} \label{equ:quant}
    l_{quantile}(a, y^Q) =  \frac{1}{N} \sum_{n=1}^N  
   \begin{cases}
        q \cdot (a_n - y^Q_n) &, y^Q_n \leq a_n \\
        (1 - q) \cdot (y^Q_n - a_n) &, y^Q_n > a_n
    \end{cases}
\end{equation}

% equation with the formula

% useless sentences
% Those variables are static in our TFT model and will be embedded into a latent vector. % The building properties are computed based on the input files of CitySim. The detailed variables could be referred to Table \ref{tab:bud_variables}. To conduct the simulation in CitySim, TMY files are required The building geometry is provided 
% The work implements the temporal fusion transformer

%% file: contents/result.tex
\section{Result \& Discussion}
Model projections will be compared with the conjectural heating and cooling demands by CitySim to quantify the effectiveness of our proposed methods. 4 additional simulations by CitySim are conducted with weather files in different climate zones and the same campus environment to produce conceptual heating and cooling demands. To assess the effectiveness of the proposed method, a classic recurrent neural network and a transformer are trained to compare with the TFT model. All models are trained for 400 epochs using AdamW Optimizer in learning rate as 1e-4. 

\begin{figure}[ht]
    \centering
    \includegraphics[width=\textwidth]{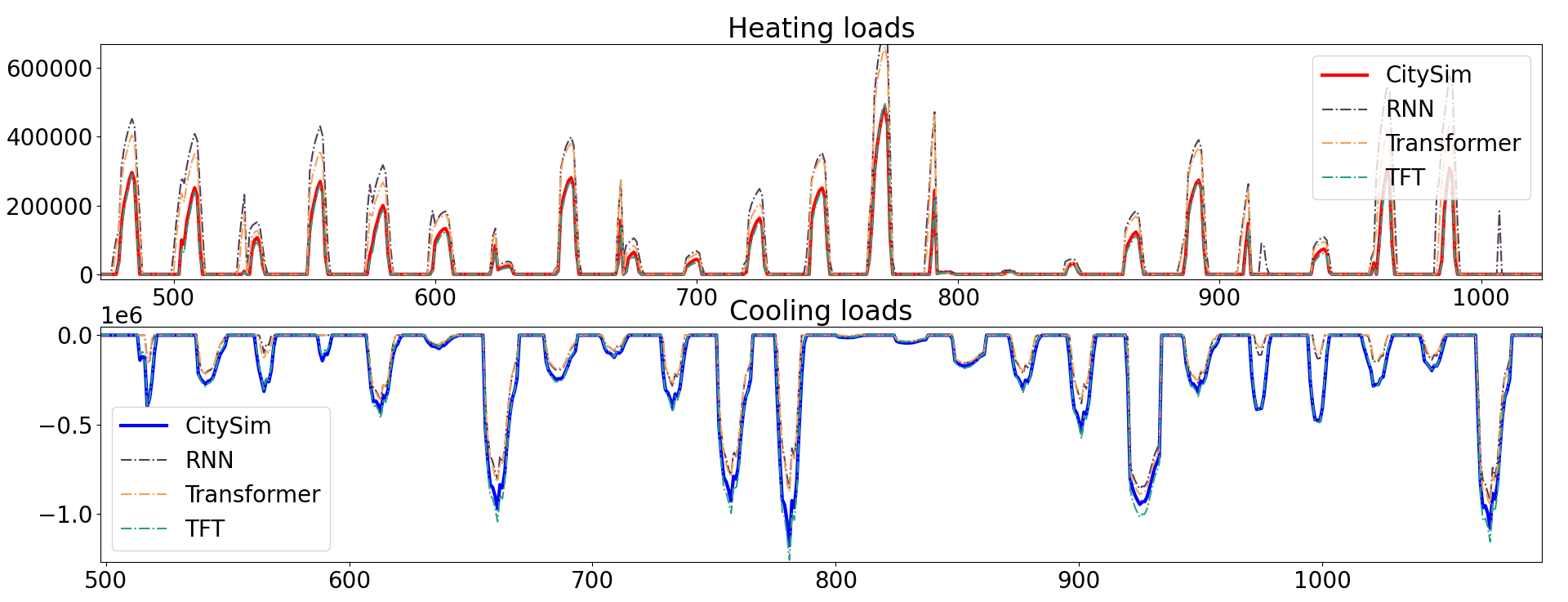}
    \caption{Comparison between ideal and predicted heating/cooling loads}
    \label{fig:pred_demo}
\end{figure}

\textbf{Ability to simulate unseen climate dynamics} 
The probabilistic and deterministic outputs from the TFT model are compared with the CitySim simulation results. Extra 4 simulations are conducted to evaluate the ability of our CityTFT model in unseen climate dynamics. The usage of climate files could be referred to Table \ref{tab:cli_files}. The prediction from CityTFT is made by observing probabilistic and quantile output at the same time. If the probability at a certain time step is over the threshold, then the prediction at the same time step is filled with the median projection in the quantile output. Fig. \ref{fig:pred_demo} demonstrates the predicted loads and ideal loads by CityTFT and CitySim. Each of the three models demonstrates a proficient ability to estimate both heating and cooling demands concluded in Table. \ref{tab:performance_overall}. In particular, CityTFT exhibits the highest level of proficiency in accurately predicting energy demands, whereas the other two models occasionally display tendencies to underestimate or overestimate the regressed values. 

\textbf{Ability to anticipate the trigger of heating and cooling.}
 To evaluate the probabilistic prediction, we compare the F1 score of three different models while classifying the CitySim results into a binary class as zero and non-zero. Table \ref{tab:performance_overall} indicates that all three models have a great ability to predict whether heating and cooling are triggered. We could also see that the scores in cooling loads are normally lower than those in heating loads shown in Appendix \ref{heating_cooling_split}. The reason is that, in physic-based UBEM tools, triggering heating systems is more straightforward which related to weather dynamics. In contrast, the activation of cooling systems is intricately affected by weather, solar heating, and human comfort examination.

 However, even though the transformer model is predicting slightly worse than the other two, it shows comparable output in Fig. \ref{fig:pred_demo} and lower MAPE values in Table \ref{tab:performance_overall}. We could observe that the predictions from the transformer are even closer than those from RNN. RNN, on the other hand, could better predict the triggering probability while the regressed projection is worse. Those findings could be observed in the demonstration in Fig. \ref{fig:pred_demo}. The debate between the self-attention and recurrent mechanisms is worth investigating more deeply. The findings suggest that employing a hybrid structure, such as CityTFT, can yield exceptional performance in this task when contrasted with the exclusive utilization of either RNN or self-attention models. 

% there should be an F1 table, an overall loads difference table
% a table compare with only non-zero loads.
% there should be a figure compare actual loads and predicted loads.
% Future Work
%This argues some improvement can be made in the future on what kinds of data TFT needs to precisely know when to trigger. 
\begin{table}[ht]
    \centering
    \begin{tabular}{ccccc}
        ~ & {\small F1 score (\%)} & {\small Non-zero RMSE (kWh)} & {\small Total RMSE (kWh)}  & {\small Non-zero MAPE} (\%)\\ \hline
        \small RNN & 91.91 & 114.06 & 75.91 & 136.89\\ 
        \small Transformer & 91.33 & 118.43 & 79.74 & 113.65\\ 
        \small \textbf{TFT} & \textbf{99.98} & \textbf{21.34} & \textbf{13.57} & \textbf{11.62}\\ 
    \end{tabular}
    \caption{F1 score, RMSE with total loads and non-zero loads Comparison}
     \label{tab:performance_overall}
\end{table}

%% file: contents/conclusion.tex
\section{Conclusion}
This work proposes CityTFT, a temporal fusion transformer, to model the urban building energy in unseen climate dynamics. This reduces the barrier for individuals to possess a meticulously designed building geometry for simulating energy demand, thereby accelerating the more precise quantification of energy modeling within the context of climate change. This permits decision-makers to conduct a thorough exploration of energy usage across multiple buildings and climate zones during the design phase, with reduced reliance on intensive simulation efforts. From a technical perspective, several facets merit additional assessment. A thorough examination should be made to evaluate CityTFT on climate change data like CMIP6. An ablation analysis involving building parameters and weather data could be additionally integrated to assess the lower limit of CityTFT's performance. The community could realize substantial benefits in the event that CityTFT could operate comparably on more concise input from publicly available data sources, such as satellite observations or open street map data.

%% file: contents/appendix.tex
\section{Appendix}
\subsection{Input Variables in CitySim and CityTFT}
Table \ref{tab:bud_variables} concludes the processed input for CityTFT. All the parameters in this table could be traced back to the geometries or some material paramters in the original CitySim XML file input.

\begin{table}[!h]
    \parbox{.40\linewidth}{
        \centering
        \begin{tabular}{|c|}
        \hline
            Building Variables                             \\ \hline
            Building height                      \\ 
            Building perimeter                   \\ 
            Wall glazing ratio                   \\ 
            Footprint area                       \\ 
            Heating setpoint temperature     \\ 
            Cooling setpoint temperature     \\ 
            Average walls U-value                \\ 
            Roof U-value                         \\ 
            First floor U-value                  \\ 
            Average windows U-value (W/m2k)      \\ 
            Average walls short-wave reflectance \\ 
            Wall short-wave reflectance          \\ 
            Roof short-wave reflectance          \\ \hline
        \end{tabular}
        \caption{Building Properties used in the training dataset}
        \label{tab:bud_variables}
    }
    \hfill
    \parbox{.45\linewidth}{
        \centering
        \begin{tabular}{|c|}
    \hline
     Weather  Variables \\ \hline
        Day of month \\
        Month\\
        Hour \\
        Diffuse radiation horizontal \\
        Beam \\
        Temperature \\
        Surface temperature \\
        Wind speed \\
        Wind direction \\
        Relative humidity \\
        Precipitation \\
        Cloud cover fraction \\
        \hline
    \end{tabular}
        \caption{Used Weather variables in TMY filest}
        \label{tab:weather_vars}
        }
\end{table}

\subsection{Distribution of loads: With or Without zero}
\begin{figure}[!htbp]
     \centering
     \begin{subfigure}{\textwidth}
         \centering
         \includegraphics[width=\textwidth]{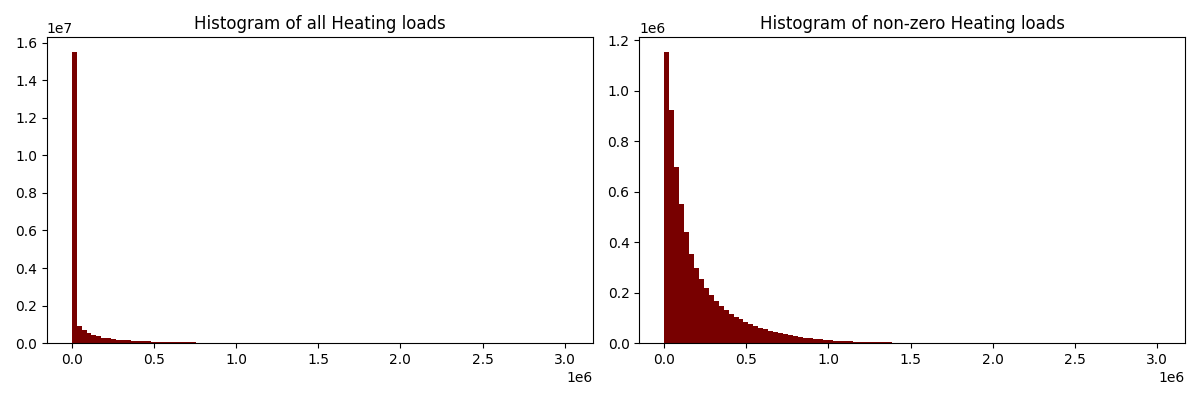}
         \caption{Distribution of heating loads}
         \label{fig:heating_hist}
     \end{subfigure}
     \hfill
     \begin{subfigure}{\textwidth}
         \centering
         \includegraphics[width=\textwidth]{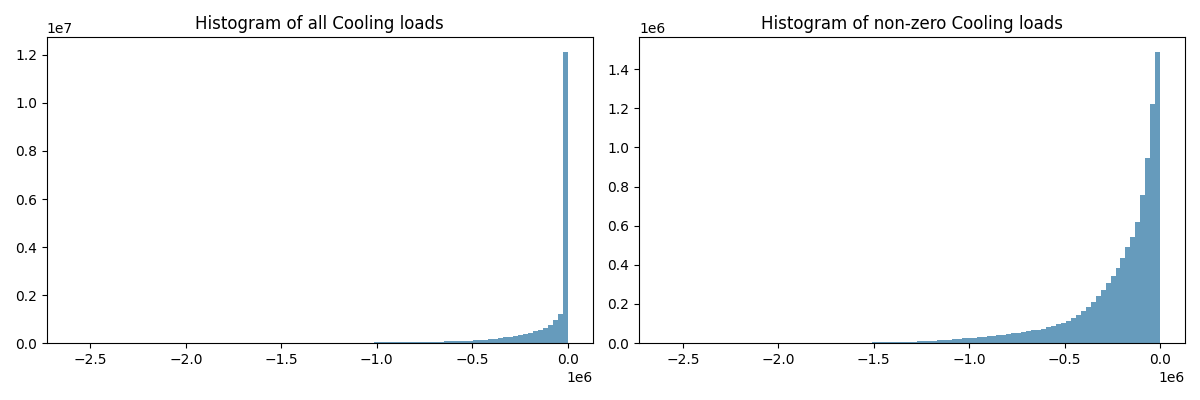}
         \caption{Distribution of cooling loads}
         \label{fig:cooling_hist}
     \end{subfigure}
        \caption{Distribution comparison between all loads and non-zero loads.}
        \label{fig:historgrams}
\end{figure}

Fig. \ref{fig:historgrams} indicate the distributions of heating and cooling load with and without zero loads. The upper two histograms are for heating loads while the lower two are suggesting cooling loads. Both heating and cooling distributions show less skew after trimming out zero values. In general, the model should be relatively easy to fit with the target distribution when the skewness is lower. The y-scale is also different in that the first bar in loads with zero demonstrates dominant height as the major distribution. This implies that when only employing regression loss to optimize the model, there is a large likelihood of the model becoming overly responsive to input variables with minimal values, consequently leading to the generation of predictions that systematically underestimate the overall energy demands. 

\subsection{Climate data collection \& Splitting datasets}
Table \ref{tab:cli_files} lists all the climate files used in this study.

\subsection{Result metrics with only heating or cooling loads} \label{heating_cooling_split}
While calculating the overall performance in the previous section, separate metrics for heating and cooling loads are also computed and shown in Table \ref{tab:performance_heat_only} and \ref{tab:performance_cool_only}. Each of those models performs better in predicting heating loads than cooling, no matter in predicting triggering or in quantile regression. 

\begin{table}[!h]
    \centering
    \begin{tabular}{ccccc}
        ~ & {\small F1 score (\%)} & {\small Non-zero RMSE (kWh)} & {\small Total RMSE (kWh)}  & {\small Non-zero MAPE} (\%)\\ \hline
        \small RNN & 93.22 & 110.55 & 64.95 & 287.37 \\ 
        \small Transformer & 93.05 & 95.19 & 61.30 & 205.88\\ 
        \small \textbf{TFT} & \textbf{99.99} & \textbf{18.24} & \textbf{9.46} & \textbf{12.92}\\ 
    \end{tabular}
    \caption{Heat only result}
    \label{tab:performance_heat_only}
\end{table}

\begin{table}[!h]
    \centering
    \begin{tabular}{ccccc}
        ~ & {\small F1 score (\%)} & {\small Non-zero RMSE (kWh)} & {\small Total RMSE (kWh)}  & {\small Non-zero MAPE} (\%)\\ \hline
        \small RNN & 90.10 & 115.78 & 85.48 & 61.86\\ 
        \small Transformer & 88.96 & 12.845 & 94.66 & 67.66\\ 
        \small \textbf{TFT} & \textbf{99.97} & \textbf{22.73} & \textbf{16.70} & \textbf{10.97}\\ 
    \end{tabular}
    \caption{Cool only result}
    \label{tab:performance_cool_only}
\end{table}

\begin{table}[h]
    \centering
\rotatebox{90}{
    \begin{tabular}{ccccccc}
    \hline
        \textbf{Location} & Lat & Lon & Elevation & Koppen Climate Zone & Climate Full Name & Train/Val/Test \\ \hline
        \textbf{Bogota/El-Dorado} & 4.7 & -74.133 & 2547.0 & C & Oceanic Climate (Warm Summer) & Train \\ 
        \textbf{Colombo} & 6.9 & 79.867 & 7.0 & A & Tropical Monsoon Climate & Val \\ 
        \textbf{Dublin Airport} & 53.433 & -6.233 & 82.0 & C & Oceanic Climate (Warm Summer) & Val \\ 
        \textbf{Jacksonville Airp. FL} & 30.5 & -81.7 & 9.0 & C & Subtropical Humid Climate (Hot Summer) & Train \\ 
        \textbf{Key West FL} & 24.55 & -81.75 & 1.0 & A & Tropical Wet And Dry Climate (Winter Dry Season) & Train \\ 
        \textbf{Kinloss} & 57.65 & -3.567 & 5.0 & C & Oceanic Climate (Warm Summer) & Train \\ 
        \textbf{Kota Bahru} & 6.167 & 102.283 & 5.0 & A & Tropical Monsoon Climate & Test \\ 
        \textbf{Marseille} & 43.433 & 5.217 & 6.0 & C & Mediterranean Climate (Hot Summer) & Train \\ 
        \textbf{Matamoros Intl} & 25.76 & -97.53 & 8.0 & B & Hot Semi-Arid Climate (Steppe) & Train \\ 
        \textbf{Medford/Jackson Co.} & 42.367 & -122.867 & 405.0 & C & Mediterranean Climate (Warm Summer) & Train \\ 
        \textbf{Mobile AL} & 30.683 & -88.25 & 67.0 & C & Subtropical Humid Climate (Hot Summer) & Train \\ 
        \textbf{Ocotal} & 13.617 & -86.467 & 612.0 & A & Tropical Monsoon Climate & Train \\ 
        \textbf{Penang/Bayan L.} & 5.3 & 100.267 & 3.0 & A & Tropical Monsoon Climate & Train \\ 
        \textbf{Portland OR} & 45.567 & -122.717 & 12.0 & C & Mediterranean Climate (Warm Summer) & Train \\ 
        \textbf{Redding CA} & 40.5 & -122.3 & 153.0 & C & Mediterranean Climate (Hot Summer) & Test \\ 
        \textbf{Reykjavik} & 64.133 & -21.9 & 66.0 & C & Subpolar Oceanic Climate (Cool Summer) & Train \\ 
        \textbf{Sacramento Airp. CA} & 38.517 & -121.5 & 7.0 & C & Mediterranean Climate (Hot Summer) & Test \\ 
        \textbf{Tucson AZ} & 32.117 & -110.933 & 779.0 & B & Hot Semi-Arid Climate (Steppe) & Test \\ 
        \textbf{Tunis} & 36.833 & 10.233 & 3.0 & C & Mediterranean Climate (Hot Summer) & Train \\ 
        \textbf{Victoria Airp. TX} & 28.85 & -96.917 & 32.0 & C & Subtropical Humid Climate (Hot Summer) & Val \\ 
        \textbf{West Palmbeach Airp.} & 26.683 & -80.117 & 5.0 & A & Tropical Wet And Dry Climate (Winter Dry Season) & Train \\ \hline
    \end{tabular}
    }
 \caption{Location information of TMY climate file for CitySim and CityTFT}
 \label{tab:cli_files}
\end{table}